# Devanagari Handwritten Character Recognition using Convolutional Neural Network

Diksha Mehta[a1], Prateek Mehta[b*]

[a]Computer Engineering, Malviya National Institute of Technology, Jaipur, Rajasthan, India
[b]Electrical and Computer Engineering, Birla Institute of Technology, Mesra, Ranchi, India

**ABSTRACT:**

Handwritten character recognition is getting popular among researchers because of its possible applications in facilitating technological search engines, social media, recommender systems, etc. The Devanagari script is one of the oldest language scripts in India that does not have proper digitization tools. With the advancement of computing and technology, the task of this research is to extract handwritten Hindi characters from an image of Devanagari script with an automated approach to save time and obsolete data. In this paper, we present a technique to recognize handwritten Devanagari characters using two deep convolutional neural network layers. This work employs a methodology that is useful to enhance the recognition rate and configures a convolutional neural network for effective Devanagari handwritten text recognition (DHTR). This approach uses the Devanagari handwritten character dataset (DHCD), an open dataset with 36 classes of Devanagari characters. Each of these classes has 1700 images for training and testing purposes. This approach obtains promising results in terms of accuracy by achieving 96.36% accuracy in testing and 99.55% in training time.

INTRODUCTION
Several computer vision issues, such as optical character recognition, license plate identification, etc., heavily rely on character categorization. A growing requirement for digitalizing handwritten Hindi documents that employ Devanagari characters is the creation of a recognition system. For Devanagari characters, optical character recognition systems have received the least attention. A few methods for segmenting and recognizing Devanagari characters are presented in [1][2]. Our work is difficult since we must deal with both dataset preparation and categorization. The Devanagari Handwritten Character Dataset (DHCD), which contains 61 thousand pictures of 36 Devanagari characters, is thus introduced in this study. Then, to categorize the characters in DHCD, we also provide a deep learning architecture. In many classification problems in computer vision, the introduction of multilayer perceptron networks has been a turning point [3]. However, choosing well-represented features has always had a significant impact on how well a network performs [4][5].

On the other hand, Deep Neural Networks don't need any features to be explicitly defined; instead, they work with the raw pixel data to produce the best features and then employ them to divide the inputs into several classes [6].

Deep neural networks have a very high number of connections and trainable parameters since they include numerous nonlinear hidden layers. Such networks are extremely difficult to train and, in order to avoid overfitting, need a very large number of instances. Convolutional Neural Network (CNN) [7] is a subclass of Deep Neural Networks that has a relatively limited number of parameters and is simpler to train. By adjusting the number of hidden layers and the trainable parameters in each layer, as well as by making accurate assumptions about the nature of the pictures, it is possible to alter how well CNN models the input dataset [8]. They may simulate intricate non-linear relationships between input and output, much like a traditional feed forward network.

However, compared to a completely linked feed-forward network of equal depth, CNN has fewer trainable parameters. Local receptive field, weight replication, and temporal subsampling are concepts introduced by CNNs and offer some degree of shift and distortion invariance [9]. Between the input and output layers, CNNs for image processing typically consist of several convolution and subsampling layers. Fully linked layers are added after these levels to produce a unique interpretation of the input data. Also, CNNs have been applied to voice recognition in addition to image recognition [10][11].

**EXPERIMENTAL SETUP**

A. DEVANAGARI HANDWRITTEN CHARACTER DATASET (DHCD)

Left-to-right abugida Devanagari, often known as Nagari, is based on the historic Brahmi script used in the Indian subcontinent. It was created in ancient India between the first and fourth centuries CE, and by the seventh century CE, it was in widespread use. The Devanagari script, used for more than 120 languages, is the world's fourth most extensively used writing system. It consists of 47 main letters, comprising 14 vowels and 33 consonants.

1) DHCD Preparation

The 61 thousand DHCD graphics [12] were created by imagining the characters written by several authors, resulting in a great deal of variance in how each character is written. We carefully clipped each character after scanning hundreds of pages with handwriting from various authors. The dataset includes only original images. The training set comprises 80% of the dataset, while the testing set comprises 20%. 12,240 images comprise the testing set, and 48,960 images constitute the training set. Figure 1 shows the handwritten characters in Acharya et al.'s [12] dataset. The writers scanned handwritten texts and meticulously clipped each character [13].

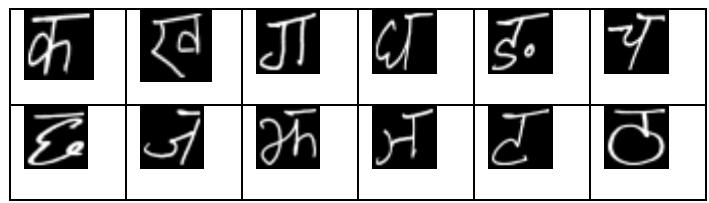

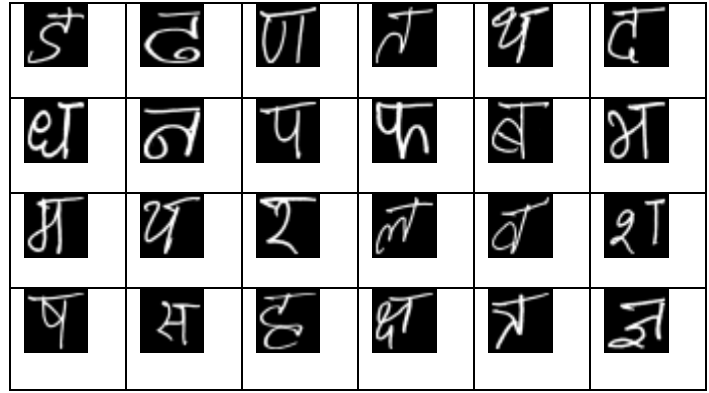

Fig. 1: Devanagari Alphabets

2) Preprocessing of the dataset

The real character is contained inside a 28x28 pixel area, and each image is 32x32 pixels in size. A padding of 2 pixels along the sides was used in the images [13]. Before the images in the dataset were padded, the cropped images underwent preprocessing. The images were initially converted to grayscale. After then, the pixels' intensities were reversed, resulting in a white letter on a dark backdrop. We repressed the background to 0 rate pixels in order to make the backdrop consistent across all of the images.

3) Obstacles faced in DHCD

Several pairs of characters in the Devanagari script found in DHCD have a comparable structure but are distinguished from one another by features such as dots and horizontal lines. As a result of how characters are written, the difficulty of the issue grows. When collecting data for DHCD, a situation resembling this was seen.

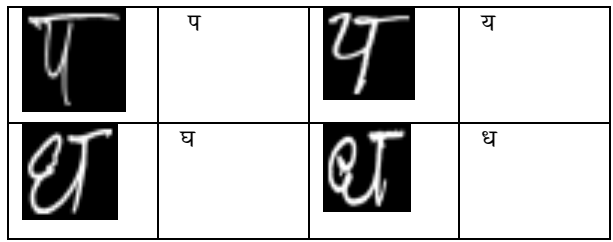

Fig. 2: Similar looking different characters

## B. CONVOLUTIONAL NEURAL NETWORK

Due to its high degree of performance across several forms of data, deep learning is quickly becoming a very useful subset of machine learning. Convolutional neural networks can be built for deep learning to classify photos (CNN). Making a CNN is simple because to the Python CV2 and Keras libraries [14].

Pixels are used by computers to view images. In photos, pixels are frequently related. An image's pattern or edge, for instance, can be represented by a particular group of pixels. Convolutions employ this to help with picture recognition [14].

Following the multiplication of a pixel matrix by a filter matrix, or "kernel," a convolution adds up the multiplication values. When every pixel in the image has been processed, the convolution moves on to the next one and repeats the procedure. CNN consists of the subsequent layers:

### 1) Convolutional Layer

The foundation of CNN is the convolution layer, which lowers the trainable weight parameters and enables parameter sharing in a single feature map. Over the input volume, the filter slides. It projects the scalar result of the dot product of all the overlapping pixels into output volume on a single slide. The input image's channel count and the filter's channel count must match.

Using this layer is best for two reasons. The first is parameter sharing, while the second is connection sparsity. The training of such a large neural network would be impossible if we just applied completely linked layers [16].

The number of parameters to learn if we apply a fully connected layer to the volumes is 32*32*3*28*28*6, which is approximately equivalent to training 14 million features. This is too many parameters to learn, and it gets much greater if we use a deeper neural network with many layers. However, if we use CNN in the same model, the number of parameters to learn would be (5*5+1) *6 = 156, which looks doable to train [16].

The parameter sharing feature of convolution is quite helpful since it allows us to employ edge detectors from earlier layers in subsequent layers as well. As a result, when a single filter is convolved, its feature is shared over the whole volume [16].

When the f * f * c filter is applied to an image of h * w * c, the output volume will have the dimensions specified in Equation (1).

$$\left\lfloor \frac{h-f+2*p}{s} + 1 \right\rfloor * \left\lfloor \frac{w-f+2*p}{s} + 1 \right\rfloor * n_f \qquad (1)$$

where,

$n_f$ = number of filters used,
s = stride, and
p = padding.

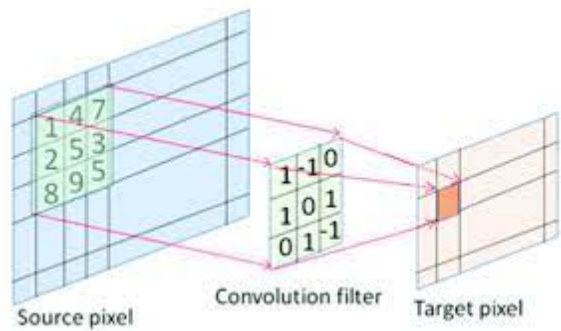

Fig. 3. Convolutional Operation Visualization

2) Rectification Linear Unit (RELU)

It is an activation function. It applies elementwise and takes 0 at the threshold value; its function is max (0, x).

3) Pooling Layer

Down sampler is all that the pooling layer is. CNN has utilized it to make representations smaller and easier to manage by reducing the size of the images. The benefit of pooling layers is that they reduce computation and guard against overfitting [15]. The Pooling layer's operation may be summed up as follows:

Accepts a volume of size W x H x D

1. Requires three parameters:
   • Spatial extent SE,
   • Stride SD
2. Produces volume of size W2 x H2 x D2 where:
   • W2 = (W1 - SE)/SD +1
   • H2 = (H1 - SE)/SD +1
   • D2 = D1
3. It introduces zero parameters because it computes a fixed function of the input.

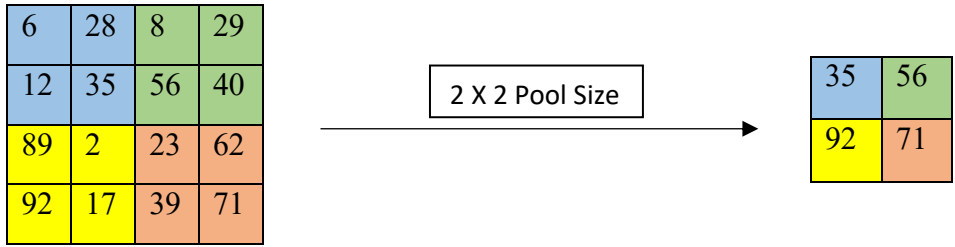

Fig. 4. Before and After Max-Pooling

4) Fully Connected Layer

It is composed of neurons that link the whole input volume, similar to a regular neural network.

C. MODEL OVERVIEW AND ARCHITECTURE

A basic convolutional neural network like the one used in our recognition approach is shown in Fig. 5. The 32x32 grayscale image's raw pixel values make up the input layer, which has no trainable parameters. The first convolution layer (C1) feature maps each contain 784 neurons (28 x 28). The graphic shows each feature map as a 2D plane with its own set of weights. Because each unit in a feature map has the same set of weights, the same features can trigger them in different locations.

In addition to providing invariance against local changes in feature position, this weight sharing significantly reduces the actual number of trainable parameters at each layer. Each layer's units receive input from a nearby neighborhood that is situated in the same area as the one below it. As a result, the neighborhood size of the preceding layer that is transferred to each unit in a convolutional layer determines how many trainable weights are associated with that unit. Since each unit is only activated by information from a certain neighborhood, it is possible for it to recognize local details like corners, edges, and endpoints. Hubel and Wiesel's studies on the locally selective, orientation-sensitive neurons in the cat's visual system [12] served as inspiration for this concept of the local receptive field.

For a 5x5 kernel, there are 25 input weights for each unit, as shown in Figure 5. The units also have a bias that is trainable. The size of the kernel in the layer underneath it and the degree of overlap between kernels define the overall number of units in that layer.

A subsampling/pooling (S1) layer comes after the convolutional layer. By averaging or pooling the neighborhood features, the sub-sampling layer reduces the resolution of the feature map produced by the convolution layer. The system shouldn't learn the absolute position of a feature but rather the relative position of features because the precise position of features varies across subsequent photos of the same character. The pooling layer, which increases the classifier's resistance to shift and distortion, helps with this goal. It gathers data from several tiny local locations and constructs a pooled feature map from that data. As a result, the local area of the previous convolution layer that provides input to the pooling layer's units determines how many units are present in the pooling layer. The dimension of the feature maps is thus decreased to half the size of the convolution layer assuming a non-overlapping method and a 2X2 region from the previous layer connected to units in the pooling layer. By identifying the greatest value in its local receptive field, multiplying it by a trainable coefficient, and adding a trainable bias, the max pooling technique generates output [12].

The second convolution layer follows the subsampling layer (C2). Layer C2's feature maps are created utilizing S1 as input. A portion of the layers in S1 and the units in C2 both get input from the 5x5 neighborhood at the same time. To reduce the overall number of trainable parameters and introduce randomization in the input distribution to various feature maps, which will allow them to learn complementary features from one another, not all C2 feature maps are linked to all S1 feature maps. The fully linked layer receives the output of this convolution layer after it has been subsampled and convolved [12]. A 1D feature vector is obtained at this stage.

The completely linked layers employ nonlinearity to represent the input, just like in a traditional feed-forward network. The nonlinearity used is known as a Rectified Linear Unit (ReLU) nonlinearity and is indicated by the symbol $f(x) = \max(0, x)$. Gradient-descent training is employed in place of frequently used non-linear functions like $f(x) = \tanh(x)$ and $f(x) = (1 + e^x)^{-1}$ because it trains ReLU much more quickly than other non-linearities like $f(x) = \tanh(x)$ and $f(x) = (1 + e^x)^{-1}$ [17].

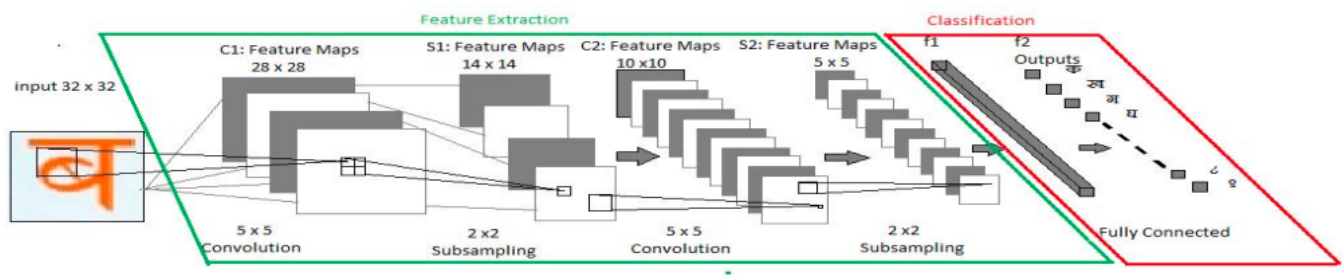

Fig. 5. Convolutional Neural Network  [12]

The network's depth and the number of its various layers are heavily dependent on the dataset and the issue area. In addition, the count of feature mappings in a layer, the size of the kernel on every layer, the selection of a non-overlapping or overlapping kernel, as well as the degree of overlap are all distinct yield outcomes [12]. Consequently, in our scenario, we evaluated many designs by modifying these parameters and reported the architecture with the greatest accuracy on the test data set. The test findings are presented in the section titled "Results and Discussion".

## RESULTS AND DISCUSSION

The presented model has been trained using 48,960 images. The dataset was split into 80:20, where 80% is for the training set, and 20% comprises the testing set. Training results achieved an accuracy of 0.9955, loss of 0.0145, precision of 0.9957, and recall of 0.9954, whereas testing results had an accuracy of 0.9636, loss of 0.2102, precision of 0.9643, and recall of 0.9632. We carried out the training till 30 epochs and noted the accuracy at each epoch.

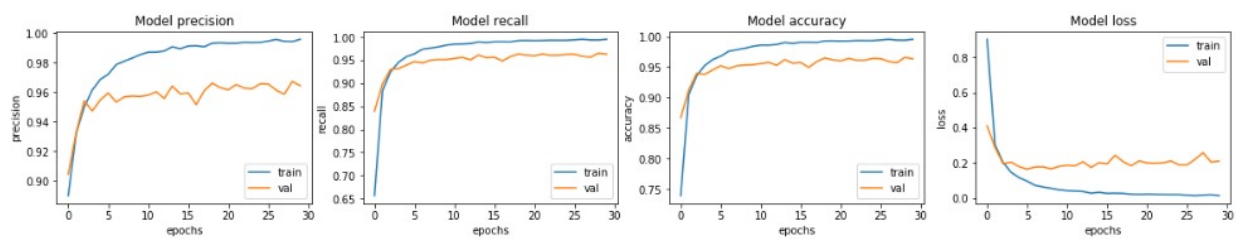

Fig. 6. Training and Testing results

## CONCLUSION

In this paper, we have presented a model having basis of a deep convolution neural network for recognition of handwritten text in Hindi with the help of Keras and Python CV2. In the experiment, the model performed very well with the given datasets. The model was trained using 48,960 images, obtaining a validating accuracy of 0.9636. The network's depth and the number of its various layers are heavily dependent on the dataset and the issue domain. There is always a scope for improvement in the machine learning models; this model can be improved by training more datasets in the future.